\documentclass{article}

\usepackage{microtype}
\usepackage{graphicx}
\usepackage{subfigure}
\usepackage{booktabs} 

\usepackage{hyperref}
\usepackage{multirow}

\usepackage[accepted]{icml2025}

\usepackage{amsmath}
\usepackage{amssymb}
\usepackage{mathtools}
\usepackage{amsthm}

\usepackage[capitalize,noabbrev]{cleveref}

\theoremstyle{plain}
\newtheorem{theorem}{Theorem}[section]

\newtheorem{lemma}[theorem]{Lemma}

\theoremstyle{definition}

\theoremstyle{remark}

\usepackage[textsize=tiny]{todonotes}

\begin{document}

\twocolumn[
\icmltitle{Free-Knots Kolmogorov-Arnold Network: On the Analysis of Spline Knots and Advancing Stability}



\icmlsetsymbol{equal}{*}

\begin{icmlauthorlist}
\icmlauthor{Liangwei Nathan Zheng}{uoa}
\icmlauthor{Wei Emma Zhang}{uoa}
\icmlauthor{Lin Yue}{uoa}
\icmlauthor{Miao Xu}{uq}
\icmlauthor{Olaf Maennel}{uoa}
\icmlauthor{Weitong Chen}{uoa}
\icmlcorrespondingauthor{Weitong Chen}{weitong.chen@adelaide.edu.au}
\end{icmlauthorlist}

\icmlaffiliation{uoa}{University of Adelaide, Adelaide, Australia}
\icmlaffiliation{uq}{University of Queensland,Brisbane, Australia}


\icmlkeywords{Machine Learning, ICML}

\vskip 0.3in
]



\printAffiliationsAndNotice{\icmlEqualContribution} 

\begin{abstract}
Kolmogorov-Arnold Neural Networks (KANs) have gained significant attention in the machine learning community. However, their implementation often suffers from poor training stability and heavy trainable parameter. Furthermore, there is limited understanding of the behavior of the learned activation functions derived from B-splines. In this work, we analyze the behavior of KANs through the lens of spline knots and derive the lower and upper bound for the number of knots in B-spline-based KANs. To address existing limitations, we propose a novel Free Knots KAN that enhances the performance of the original KAN while reducing the number of trainable parameters to match the trainable parameter scale of standard Multi-Layer Perceptrons (MLPs). Additionally, we introduce new a training strategy to ensure $C^2$ continuity of the learnable spline, resulting in smoother activation compared to the original KAN and improve the training stability by range expansion. The proposed method is comprehensively evaluated on 8 datasets spanning various domains, including image, text, time series, multimodal, and function approximation tasks. The promising results demonstrates the feasibility of KAN-based network and the effectiveness of proposed method. Our code is openly available: \url{https://github.com/IcurasLW/FR-KAN.git}.
\end{abstract}

\section{Introduction}

Multi-Layer Perception(MLP) is the dominated architecture in current deep network, which is a well-known as piecewise linear spline operator $f_{\Theta}: \mathbb{R}^q \rightarrow \mathbb{R}^p$ that maps an input signal $\mathbf{x} \in \mathbb{R}^q$ to an output $\mathbf{y} \in \mathbb{R}^p$ \cite{balestriero2018mad,balestriero2018spline,telgarsky2016benefits}. The learning process aligns with Universal Approximation Theorem(URT) \cite{hornik1989multilayer} as each neuron contribute one knot to the final piecewise linear spline at a learnable position. This power enables MLP to approximate any continuous function by sufficient number of neurons. Recently, Kolmogorov–Arnold Networks(KANs) \cite{liu2024kan} has attracted significant attention of the community due to high interpretability and the inspiring learning paradigm. Unlike MLP, KANs follow the Kolmogorov-Arnold Representation Theorem(KAT), which states that every multivariate continuous function $f: [0,1]^n \rightarrow \mathbb{R}$ can be represented as a superposition of continuous single variable functions\cite{sprecher2002space,liu2024kan,schmidt2021kolmogorov,leni2013kolmogorov}. The architecture with B-spline KAN can be considered viewed as fixed knots spline operator where the knots are not fully learnable, limiting the expressive power of KAN. In this paper, we aim to bring a new insight to reveal behavior of KAN from spline knots perspective and analyze the bound for the spline knots of KAN. Current community consistently believe that KAN is weaker than MLP in typical deep learning task by preliminary experiment. However, our analysis and comprehensive experiments show that multilayer and shallow KAN network actually outperform MLP under the same architecture setting, shedding light on why KANs often outperform traditional Multi-Layer Perceptrons (MLPs) across various tasks.

KANs assume that the target function is the composition of smooth and uniform function, which can be expressed as typical basis function like $e^x$, $log(x)$ and $x^n$. This makes KANs well-fit on scientific task such as function approximation and symbolic regression, significantly winning MLP on AI-for-sciencne tasks. Despite this, the original implementation is trivial and still infeasible for typical deep learning tasks such as image classification. We hypothesis that the uniform grid assumption is difficult to fit finite observation dataset from real world distribution and the scalability of KAN is also one of the obstacles for common machine learning task. There are still some challenges faced by original KANs: \textbf{(1) Constrained Expressive Power}: The expressive ability of KAN is constrained by the fixed grid segment. Neural networks are known as learnable spline operator from the approximation theory perspective\cite{balestriero2018mad,balestriero2018spline,telgarsky2016benefits}. Original KAN employed learnable fixed grid B-spline function as the activation. This fixed knots setting can constraint the development of knots, resulting less expressive power of network. The fixed grid does not generate more spline knots even if more neurons are added. We show a bound of knots for the original KAN (Theorem \ref{theorem:KAN_Knots_Bound}) that the spline knots learned from the original KAN only correlated to grid size, spline order and layers of networks. This bound inspires us to unleash the power of KAN from fixed knots and bring new insights to understand the behavior of KAN. \textbf{(2) Heavy Trainable Parameter}: The original KANs learn a unique activation function for each neuron, which introduces additional parameters proportional to \( G + K \) for the spline combination compared to a standard MLP. This results in a parameter complexity of \( \mathcal{O}((d_{in} \times d_{out})(G + K + 1)) \), whereas a standard MLP has a parameter complexity of \( \mathcal{O}((d_{in} \times d_{out}) + d_{out}) \). \textbf{(3) Oscillation and Unstable Training}: The original KAN implementation constrains the activation function within a narrow range (e.g., [-1, 1]) and no discussion on the impacts of grid range, this constraint becomes problematic as the grid size increase, leading to densely packed grid points within the grid range. Consequently, this exacerbates oscillation and instability, particularly when stacking additional layers, often resulting in unstable training and NaN values in the output. Such oscillation limit the expressive power of the network by restricting the grid size and spline order. We demonstrate the underlying causes of activation oscillation through preliminary experiments shown in Figure \ref{fig:simple_visual}.

To address the above problems, we propose \textbf{FR}ee-knots \textbf{KAN} (\textbf{FR-KAN}) motivated by the bound of spline knots of KAN. FR-KAN aims to release the constraint of fixed knots and significantly reduce the trainable parameter size to MLP level by disentangling the grid vector from number of neuron. We also release the constraints causing serious oscillation by regularizing the second derivatives of spline weight and enlarge the grid range. This makes KAN network to have more grid size than the original KAN for smoother fitting and stable training. Theoretically, we show the spline knot bound of KAN only correlated to grid size $G$, spline order $K$ and network layer $L$, which provides theoretical insights for understanding behavior of KAN network with B-spline. This insight can also extend to other KAN variants. We summarize our main contributions as below:

\textbf{(C1) Knots Upper Bound:} Derive a tight bound for spline knots in KAN to understand the behavior of KAN and to identify potential factors that may impact performance.

\textbf{(C2) Free Grid Spline:} Extend KAN by replacing the fixed-grid constraint with a free-knots grid approach, enabling more adaptable spline configurations to enhance expressive power by the non-uniform grid.

\textbf{(C3) Stability Improvements:} Address the severe oscillation issues inherent in the original KAN by implementing second derivative regularization, resulting in smoother training dynamics and improved overall performance. 

\textbf{(C4) Comprehensive Evaluation:} Implement fair experiments on 9 datasets including image, text, time series, multimodal, and function approximation, showcasing superior performance compared to MLP, KAN in terms of comparable parameter and performance.


\section{Related Works}
Kolmogorov-Arnold Networks (KANs) is firstly proposed by \cite{liu2024kan} for scientific tasks such as symbolic regression. The promising performance and interpretability of KANs arouse active discussion in the machine learning community. Many researchers start to transfer KANs replace MLP in traditional network architecture such as KAN-CNN\cite{bodner2024convolutional}, Graph-KAN\cite{zhang2024graphkan}, KAN-Transformer\cite{genet2024temporal,yang2024kolmogorov}, and KAN-UNet\cite{li2024u}. In particular, \cite{yang2024kolmogorov} noticed that KAN is extremely difficult to scale up due to the additional grid size and the recursive generation of B-spline. Some researchers started to look for an alternative activation for B-spline function while some are consistent with B-spline due to high performance. Therefore, we categorized existing method into two steams: Kernel-based KAN and Spline-based KAN. Original KAN \cite{liu2024kan} is a typical Spline-based KAN that learns multiple unique spline function as the activation of network. B-spline function is the basis of original KAN where each segment of spline is recursively determined, resulting significantly increasing computational cost. 

On the other hand, Kernel-based KAN kernels has closed-formed expression without recursively computational such as RBF kernel, polynomial and rational kernel. In particular, Chebyshev-KAN \cite{ss2024chebyshev}, Rational-KAN \cite{yang2024kolmogorov}, Wavelet-KAN\cite{bozorgasl2024wav}, Fourier-KAN\cite{xu2024fourierkan}, and RBF-KAN \cite{li2024kolmogorovarnold} are the KAN network leverage the power of various kernels. The advantage of Kernel-based KAN usually has lower computational cost than original-KAN since the recursion evaluation is not required.


\section{Kolmogorov-Arnold Network Architecture}
Original Kolmogorov-Arnold Network(KAN) includes two components. One is learnable B-spline function and one is SiLU activation function. In the original implementation, the output of SiLU function is simply summed with the output of B-Spline function. Formally, given an input $x_i \in \mathbb{R}^q$ and a $K$-order B-spline with $G$ grid size, KAN's architecture can be written as:

\begin{equation}
    \phi(x_i) = A_b \sum_{j=0}^{G} c_j B_j(x_i) + A_s SiLU(x_i) \label{eq:KAN_architecture}
\end{equation}

where, $A_b$ and $A_s$ are the linear combination weights and $c_j$ are the weight of combination of each spline. The initialization of grid follows Uniform distribution $G \sim U[a, b]$, where $[a, b]$ is grid range. Intuitively, KAN holds the following distinct architecture to MLP: (1) KAN has learnable activation function with arbitrary order while activation in MLP is fixed. (2) KAN apply post linear combination $A_b, A_s$ after activation while MLP is pre-linear combination before activation. These two distinct differences ensure KAN to follow Kolmogorov-Arnold Theorem, which a complex function is superposition of continuous single variable functions.

\begin{figure*}
    \centering
    \includegraphics[width=0.9\linewidth]{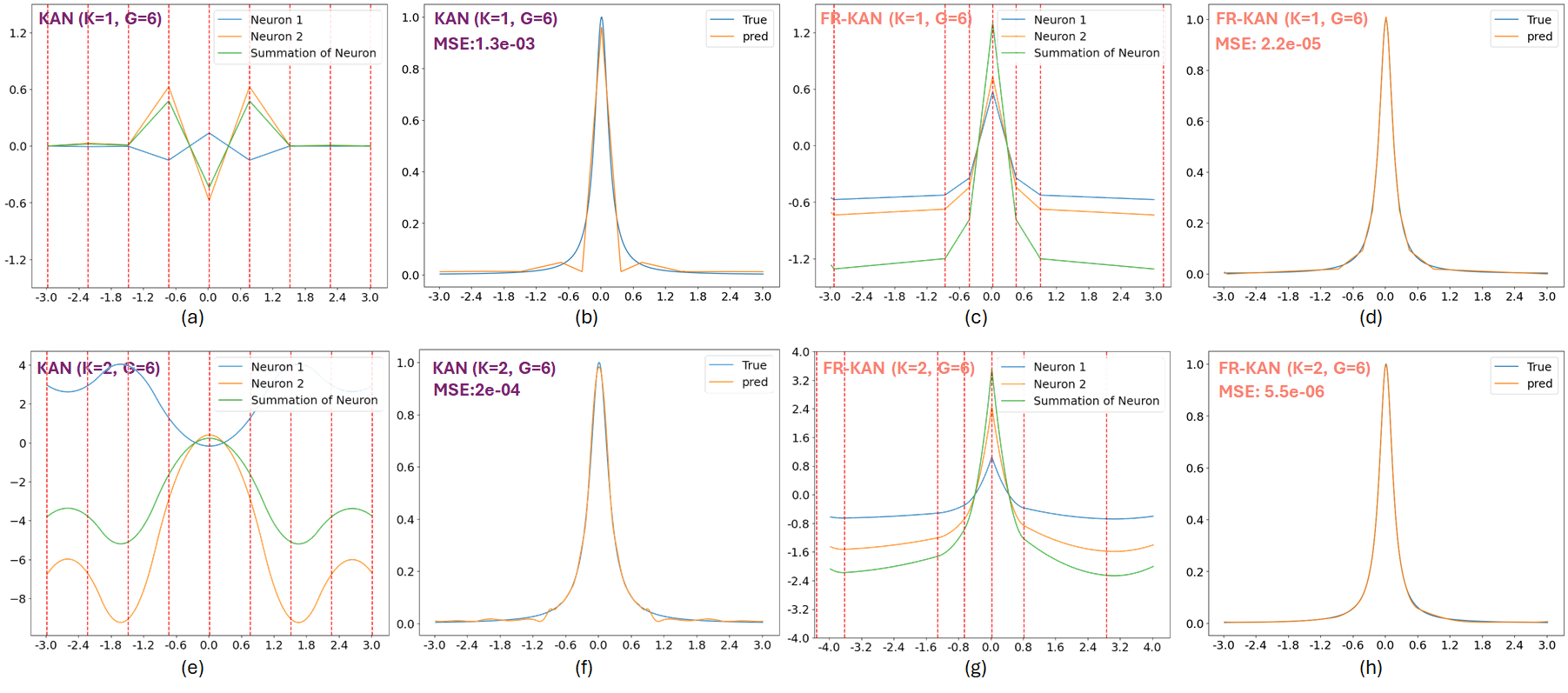}
    \caption{Preliminary Experiment: Fitting function $\frac{1}{(1+25x^2)}$, (a)(e)(c)(g): Activated feature after input layer and activation summation. (b)(d)(f)(h): Function approximation results.}
    \label{fig:simple_visual}
\end{figure*}

\section{Upper Bound of Knots for KAN}
\subsection{Number of Knots for Single Layer KAN}
Original KAN select B-spline function as the basis of learnable activation, which inherently introduce spline knots into the final spline function with the number of knots determined by the grid points of B-spline. The linear combination of basis functions in B-splines constitutes an affine transformation without affecting the grid knots, as illustrated in Figure \ref{fig:simple_visual} (a). The summation of neurons have no impacts on the position of grid points. For ease of presentation, we denoted a KAN network $\mathcal{K}_b^l \in \mathbb{R}^q \rightarrow \mathbb{R}^p$ as any $K$-order KAN network with layer $l=1,2,\dots,L$ and uniform $G$ grid size without any special boundary condition (e.g. clamped, repetitive knots), and $\mathcal{F}_r^l \in \mathbb{R}^q \rightarrow \mathbb{R}^p$ as a ReLU MLP neural network. To determine the upper bound of knots in proposed method, we first show the number of knots of a one-layer simple KAN network with B-spline functions.

\begin{lemma} \label{lemma:knot_generate_destroy}
    Any arbitrary neuron $n_i \in \mathcal{F}_r^1$ can generate up to one knot in the current spline. A knot is considered as \textbf{Unique} if the position of knot is different from other. The final knots of spline generated by $\mathcal{F}_r^1$ is up to number of neuron and depends on the number of unique knot. Therefore, any fixed grid $\mathcal{K}_b^1$ has unique $G+K$ knots since the grids are the same for all neurons.
\end{lemma}

Lemma \ref{lemma:knot_generate_destroy} is simple to prove. We know the fact that $n_i \in \mathcal{F}_r^1$ can generate a new knot if $x_i > 0$, otherwise destroy \cite{pinkus1999approximation,chen2016upper}. Because the grid of $n_i \in \mathcal{K}_b^1$ are fixed at the same position. The knots in each neuron simply vary alongside the position of the grid without shifting over it. Therefore, the linear combination of all the neurons does not generate or destroy any knots but hovers at the grid position, resulting in $G+K$ knots. The SiLU shotcut connection, on the other hand, has less contribution to the knots of final spline as the knot of SiLU activation is also fixed, but it provides simple gradient when the input falls outside of grid range. The interesting insights of this property is that, \textbf{single neuron in KAN generates $G+K$ times more knots than MLP}. Those knots provide approximation freedom to fit a target function.

\subsection{Number of Knots for Multi Layer KAN}
We know from Lemma \ref{lemma:knot_generate_destroy} that the number of knots in single layer is simply equal to $G+K$ no matter how many neurons in the layer and the shortcut connection of SiLU does not impact the number of knots in the grid range. Although one neuron in KAN generate $G+K$ more knots than MLP, but the entire network is constrained by fixed number of knots since the grids are fixed. To obtain more knots without increasing $G$ and $K$, making the network deeper is one of the common approaches. Next, we show how the knots of KAN increase by adding more layers and give the upper bound of knots of multilayer KAN. For a multilayer network, let us denote the number of knots of the previous layer by $m_{l-1}$ and the neuron of the layer by $i=1,2,\dots,n_i$.

\begin{lemma}
    \label{lemma:knots_mul_layer}
    \cite{chen2016upper} \textbf{(Spline Knots Upper Bound of ReLU-MLP)} For $L = 1$, $n_i \in F_r^l$ can generate the number of knots up to the knots of activation. For $l > 1$, $n_i \in F_r^l$ can preserve at most $m_{l-1}$ from the previous layer and generate at most unique new knots $n_i(m_{l-1}+1)$ in the current layer. Therefore, the upper bound of knots for MLP is given by
    
    \begin{equation}
        m_l \leq m_{l-1} + n_i(m_{l-1}+1)
    \end{equation}
\end{lemma}

From Lemma \ref{lemma:knots_mul_layer}, we know that $n_i$ in the network can preserve knots from the previous layer and generate new knots in the current layer. We show the proof of Lemma \ref{lemma:knots_mul_layer} in Appendix \ref{proof:Lemma_kots_mul_layer}. 

Let us denote $\triangle y_i$ as the difference of input in grid segment $\triangle g$. Given that B-spline function has at least knots as $G+K$ and Lemma \ref{lemma:knot_generate_destroy}, \ref{lemma:knots_mul_layer}. We give the upper bound of Multilayer KANs as stated in Theorem \ref{theorem:KAN_Knots_Bound}. 

\begin{theorem}
\label{theorem:KAN_Knots_Bound}
(\textbf{Spline Knots Bound of KAN}) KANs only generate knots when $\triangle y_i \geq \triangle g$, where the new knots are lying in between $[g_j, g_{j+1}]$ and the number of new knots is up to $G-1$. For any arbitrary multilayer fixed grid KAN network $\mathcal{K}_b^l$ with $K$-order and $G$ grid size B-spline function, the \textbf{Tight} bound on the number of knots $\mathcal{N}_k$ is given by:
\begin{equation}
    (G+K) \leq \mathcal{N}_k \leq (G+K) + \prod_{l=1}^l G(G-1)
\end{equation}

\begin{equation}
    \mathcal{N}_k = m_{l-1} + \sum_{i=1}^{G-1} \lfloor \frac{\Delta y_i}{\Delta g} \rfloor
\end{equation}
\end{theorem}

We show the proof of Theorem \ref{theorem:KAN_Knots_Bound} in Appendix \ref{proof:tightness_proof}. From Theorem \ref{theorem:KAN_Knots_Bound}, we understand that the number of knots of KAN completely depends on grid size $G$ and the B-spline order $K$ while MLP depends on the number of neurons. Apart from Kolmogorov-Arnold theorem, this bound indicate that the performance of KAN also depends on the number of spline knots. However, the fixed knots KAN share the same grids over all neurons, which limits the development of knots. To address this fixed knots problem and obtain higher expressive power, one feasible approach is to generate activation with unique knots for neuron $n_i$. Therefore, we proposed a Free-Knot Kolmogorov-Arnold Network to release this constraint.

\section{Free-Knot KAN}
\subsection{Neuron Grouping and Share Weight}
We know that MLP consumes trainable parameter $O(d_{in} \times d_{out} + d_{out})$ in memory. The original KAN learns a unique activation B-spline for each neuron recursively and a separate linear combination weight for the shortcut SiLU activation, resulting in $O(d_{in} \times d_{out} \times (G+K+1))$ trainable parameter size. This obstructs the scalability of KAN to large task like computer vision and NLP task. Inspired by an innovative work \cite{yang2024kolmogorov}, we employed Neuron Grouping in our proposed method to save computational cost. Specifically, we divide the $G$ of each neuron into $h$ groups, where each group share the same grid $G$ in grid range $[a,b]$. From Equation \ref{eq:KAN_architecture}, SiLU activation starts the positive activation at $x=0$ and KAN does not shift the input data $x$ as MLP does. Therefore, this SiLU activation does not introduce new knots, but remains activation value from $[b, +\infty]$. We also empirically found that $A_s$ does not significantly improve the model performance in some task such as computer vision. Therefore, we share the spline weight $A_b$ with SiLU weight $A_s$, denoted as $A$. It is worthy to notice that linear combination weight $A$ is not grouped with activation since we wish to have more variety for the activation following KAT theorem. Specifically, the number of trainable parameters becomes $O(2 \times d_{in} \times d_{out} + h(G+K))$ after the neuron grouping is applied and the trainable parameters size becomes $O(d_{in} \times d_{out} + h(G+K))$ after weight-sharing. Therefore, we decouple the trainable parameter size of KAN to the level of MLP. Equation \ref{eq:FRKAN_architecture} shows the final architecture of FR-KAN:

\begin{equation}
    \phi(x_i) = A \sum_{j=0}^{G} c_j B_j(x_i)_{j/h} + A SiLU(x_i) \label{eq:FRKAN_architecture}
\end{equation}

\begin{figure*}
    \centering
    \includegraphics[width=0.9\linewidth]{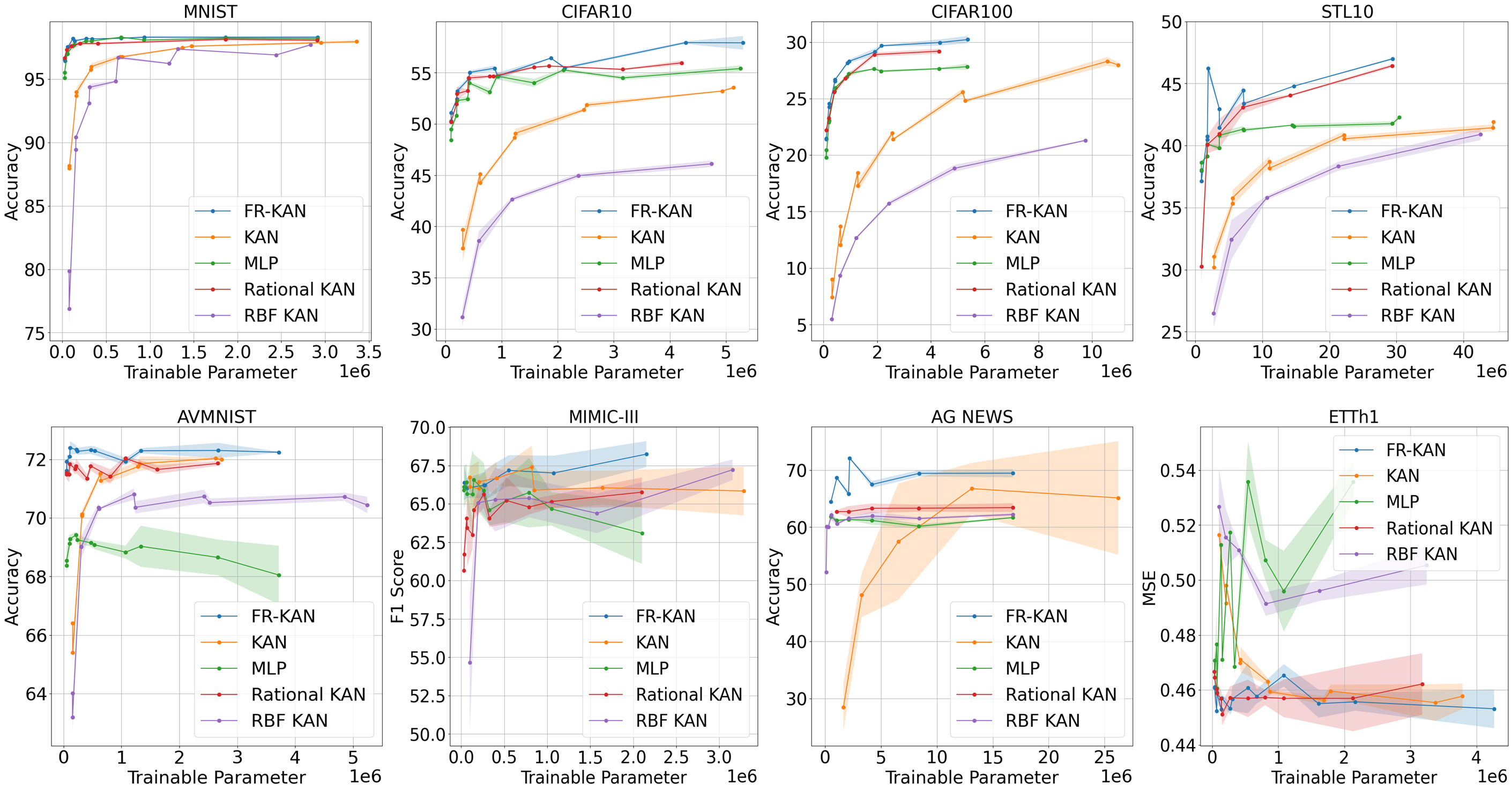}
    \caption{Performance Comparison: Row 1 is Image Classification dataset and Row 2 is Multimodal (AVMNIST, MIMIC-III), AG NEWS(Text Classification), ETTh1(Time Series Forecasting)}
    \label{fig:All_performance}
\end{figure*}

\subsection{Free Grid}
Neuron grouping is an efficient trick to reduce computational cost, but it may also reduce the performance of KAN since the neurons in the group learned the same activation. To mitigate this effect and maintain the performance from KAN, we further propose free grids to introduce more knots in the spline function. From Theorem \ref{theorem:KAN_Knots_Bound}, we propose a free-knot KAN that applies a learnable bias $b_g$ on the fixed grid $G$ defined in $[a, b]$ such that each neuron $n_i$ has a different shift on $g_j$. Ideally, we expect each neuron to have Unique shift on $G$ so that the knots are Unique. Formally, we have non-uniform grid $G^*$ as:

\begin{equation}
\label{eq:free_grid}
    G^* = Sort(G + b_g)
\end{equation}

Adding a learnable shift $b_g \in \mathbb{R}^{G}$ to grids can increase the possibility to generate more unique knots, where $b_g$ is randomly initialized from range $[-\frac{a+b}{ZG}, \frac{a+b}{ZG}]$ and optimized by gradient descent to find the best fit parameter. The $Z$ is a parameter to control the initialization range, set $Z=8$ gives us best performance. Although we can not guarantee that the knots are unique for all neurons, This minor shift can naturally introduce more knots located at different position. To avoid the case that positional reversal occurred between two grid point, that is $g_{j-1} > g_j$, and break the mathematical form of B-spline. We sorted the entire grid to maintain the grid order $g_1 < g_2 < \dots < g_{j}$. 

To this point, we unleash the fixed grid constraint to introduce more unique knots and group the neuron to reduce complexity of KAN. Suppose our free grid always generate unique knots and $\triangle y_g \geq \triangle g$ in grid range $[a,b]$ and denoted our FR-KAN network as $K_f^l$. We can obtain knots upper bound for Free-Knot KAN as stated in Theorem \ref{theorem:free-knots_KAN_Knots_Bound}:

\begin{theorem}
    \label{theorem:free-knots_KAN_Knots_Bound}
    (\textbf{Spline Knots Upper Bound of Free-Knot KAN}) For any arbitrary multilayer free grid KAN network $\mathcal{K}_f^l$ with $K$-order and $G$ grid size B-spline function, a tight upper bound on the number of knots of $\mathcal{N}_k$ is given by:
\begin{equation}
    (G+K) \leq \mathcal{N}_k \leq h(G+K) + \prod_{l=1}^l hG(G-1)
\end{equation}

\end{theorem}

\subsection{Smooth Oscillation and Training Stability}
Training a KAN network is very unstable since multilayer KAN in B-spline can easily introduce spline oscillation (inflection point) when the network goes deeper and the grid is clustered in a limited interval. The primary reason is that the nature of B-spline basis function are oscillated at inflection point. Large oscillation usually comes with large second derivative value of activation so that minor changes on input data $x_i \in R^q$ can possibly cause largely different on the amount of gradient, leading to unstable training. Although normalization (LayerNom, BatchNorm) tricks is leveraged to stabilize the training as we did in typical MLP, KAN is still unstable to train.

The cause behind this poor stability may be that the clustered grid so that the oscillation is worse. To mitigate this oscillation, we wish to ensure the smoothness of $C_2$ continuity for the activation. We propose a novel regularization term that constraint the second-derivatives of $c_j$ w.r.t input data to smooth oscillation. The main idea is to explicitly reduce inflection point and remain the activation value of B-spline. Formally, given any task objective $J(x)$ and denoted $c_{j,i}$ as spline combination weight of neuron $n_i$ at segment $g_j$, our optimization objective can be expressed as:

\begin{equation} \label{eq:loss_function}
    \mathcal{L}(x) = J(x) + \lambda \sum^{n_i}_{i=0} \sum_{j=0}^G \frac{ \partial^2 c_{j,i}}{ \partial^2 x}
\end{equation}

In addition, we further address the clustered grid by expanding the grid range of B-spline to larger interval (e.g. [-10, 10]) to relieve intensive grid cluster and provide wider non-linear receptive range. We showed the emperical results of our finding in Figure \ref{fig:Impact_Grid_Range}.

\begin{table*}[ht]
    \centering
    \resizebox{0.9\textwidth}{!}{
    \begin{tabular}{cc|cccccc}
    \textbf{Dataset} & \textbf{Hidden Layer} & \textbf{ReLU-MLP} & \textbf{KAN} & \textbf{Fourier-KAN} & \textbf{Rational-KAN} & \textbf{RBF-KAN} & \textbf{FR-KAN} \\
    \hline
    \multirow{6}{*}{\rotatebox{90}{CIFAR10}} 
        & [in, 32, out]      & 48.41 $\pm$ 0.24 & 46.02 $\pm$ 0.13 & 43.56 $\pm$ 0.39 & 50.20 $\pm$ 0.37 & 44.96 $\pm$ 0.24 & \textcolor{red}{\textbf{50.81 $\pm$ 0.27}} \\
        & [in, 64, out]      & 50.82 $\pm$ 0.19 & 47.68 $\pm$ 0.13 & 45.02 $\pm$ 0.35 & 51.94 $\pm$ 0.17 & 46.11 $\pm$ 0.32 & \textcolor{red}{\textbf{52.74 $\pm$ 0.12}} \\
        & [in, 128, out]     & 52.43 $\pm$ 0.30 & 49.38 $\pm$ 0.33 & 45.24 $\pm$ 0.71 & 53.23 $\pm$ 0.15 & 47.94 $\pm$ 0.17 & \textcolor{red}{\textbf{53.52 $\pm$ 0.33}} \\
        & [in, 32 32, out]   & 49.79 $\pm$ 0.46 & 47.46 $\pm$ 0.43 & 39.16 $\pm$ 0.41 & 50.22 $\pm$ 0.22 & 44.36 $\pm$ 0.60 & \textcolor{red}{\textbf{50.29 $\pm$ 0.02}} \\
        & [in, 64 64, out]   & 51.96 $\pm$ 0.27 & 49.86 $\pm$ 0.19 & 41.15 $\pm$ 0.23 & \textcolor{red}{\textbf{52.96 $\pm$ 0.68}} & 46.17 $\pm$ 0.03 & 52.41 $\pm$ 0.28 \\
        & [in, 128 128, out] & 54.09 $\pm$ 0.23 & 50.95 $\pm$ 0.20 & 41.86 $\pm$ 0.53 & 54.48 $\pm$ 0.31 & 47.94 $\pm$ 0.40 & \textcolor{red}{\textbf{55.03 $\pm$ 0.15}} \\
        \hline
    \multirow{6}{*}{\rotatebox{90}{STL10}} 
        & [in, 32, out]      & 37.87 $\pm$ 0.48 & 37.36 $\pm$ 0.05 & 32.00 $\pm$ 1.31 & 39.50 $\pm$ 0.55 & 38.32 $\pm$ 0.36 & \textcolor{red}{\textbf{40.21 $\pm$ 0.55}} \\
        & [in, 64, out]      & 39.29 $\pm$ 0.09 & 37.96 $\pm$ 0.08 & 32.39 $\pm$ 0.53 & 40.07 $\pm$ 0.71 & 40.90 $\pm$ 0.46 & \textcolor{red}{\textbf{42.65 $\pm$ 0.24}} \\
        & [in, 128, out]     & 39.86 $\pm$ 0.40 & 39.30 $\pm$ 0.16 & 38.55 $\pm$ 0.31 & 41.21 $\pm$ 0.75 & 42.78 $\pm$ 0.12 & \textcolor{red}{\textbf{44.02 $\pm$ 0.13}} \\
        & [in, 32 32, out]   & 38.02 $\pm$ 0.26 & 37.24 $\pm$ 0.59 & 28.28 $\pm$ 0.38 & \textcolor{red}{\textbf{39.60 $\pm$ 0.22}} & 37.76 $\pm$ 0.53 & 37.13 $\pm$ 0.10 \\
        & [in, 64 64, out]   & 39.15 $\pm$ 0.08 & 38.65 $\pm$ 0.34 & 31.58 $\pm$ 0.67 & \textcolor{red}{\textbf{42.74 $\pm$ 0.25}} & 39.84 $\pm$ 0.26 & 42.73 $\pm$ 0.11 \\
        & [in, 128 128, out] & 39.81 $\pm$ 0.23 & 40.46 $\pm$ 0.36 & 35.81 $\pm$ 0.29 & 42.05 $\pm$ 0.38 & 41.42 $\pm$ 0.49 & \textcolor{red}{\textbf{42.96 $\pm$ 0.07}} \\
        \hline
    \end{tabular}
    }
    \caption{Performance Comparison by Controlling Hidden Layer}
    \label{tab:Comparison_Hidden_Structure}
\end{table*}

\begin{table*}[ht]
    \centering
    \resizebox{0.9\textwidth}{!}{
    \begin{tabular}{c|c|cccccc}
         \textbf{Feynman Eq.} & \textbf{Original Formula} & \textbf{MLP} & \textbf{KAN} & \textbf{Fourier-KAN} & \textbf{Rational-KAN} & \textbf{RBF-KAN} & \textbf{FR-KAN} \\
         \hline
         I.6.2 & $exp(\frac{\theta^2}{2\sigma^2}) / \sqrt{2 \pi \sigma^2}$ & 1.5850 & 1.1485 & 1.4185 & 1.1162 & 1.2866 & \textcolor{red}{\textbf{0.9442}} \\
         I.6.2b & $exp(\frac{(\theta - \theta_1)^2}{2\sigma^2})/\sqrt{2 \pi \sigma^2}$ & 0.7079 & 0.6946 & 0.7191 & 0.6392 & 0.6457 & \textcolor{red}{\textbf{0.4773}} \\
         I.9.18 & $\frac{Gm_1m_2}{(x_2 - x_1)^2 + (y_2 - y_1)^2 + (z_2 - z_1)^2}$ & 0.4261 & 0.4195 & 0.4550 & 0.4178 & 0.4159 & \textcolor{red}{\textbf{0.4139}} \\
         I.12.11 & $q(E_f + B\mathcal{v}sin\theta)$ & 0.2745 & \textcolor{red}{\textbf{0.0983}} & 0.3225 & 0.0997 & 0.1170 & 0.0988 \\
         I.16.6 & $\frac{u + v}{1 + \frac{uv}{c^2}}$ & 2.3691 & 2.3665 & 2.4726 & 2.3651 & 2.3649 & \textcolor{red}{\textbf{2.3628}} \\
         I.18.4 & $\frac{m_1r_1 + m_2r_2}{m_1 + m_2}$ & 0.1606 & 0.0559 & 0.1677 & 0.1181 & 0.0763 & \textcolor{red}{\textbf{0.0521}} \\
         I.26.2 & $arcsin(nsin\theta_2)$ & 0.2394 & 9.50e-3 & 0.2453 & 0.0394 & 0.0186 & \textcolor{red}{\textbf{7.61e-3}} \\
         I.29.16 & $\sqrt{x_1^2 + x_2^2 - 2x_1 x_2 cos(\theta_1 - \theta_2)}$ & 0.2910 & 0.1337 & 0.3900 & 0.1645 & 0.1494 & \textcolor{red}{\textbf{0.1317}} \\
         I.30.3 & $I_*,0 \frac{sin^2(\frac{n\theta}{2})}{sin^2(\frac{\theta}{2})}$ & 0.1874 & 2.46e-3 & 0.2522 & 4.69e-3 & 2.47e-3 & \textcolor{red}{\textbf{2.33e-3}} \\
         I.50.26 & $x_1(cos(wt) + \alpha cos^2(wt))$ & 0.1162 & 0.0911 & 0.3397 & 0.0471 & 0.0408 & \textcolor{red}{\textbf{0.0407}} \\
         II.11.27 & $ \frac{n\alpha}{1-\frac{n\alpha}{3}}\epsilon E_f$ & 0.1082 & 0.0764 & 0.1085 & 0.1063 & 0.0873 & \textcolor{red}{\textbf{0.0650}} \\
         II.35.18 & $\frac{n_0}{exp(\frac{\mu m B}{k_b T} + exp(-\frac{\mu m B}{k_b T}))}$ & 0.1692 & 0.0896 & 0.1949 & 0.0972 & 0.0798 & \textcolor{red}{\textbf{0.0603}} \\
         III.10.19 & $\mu_m \sqrt{B_x^2 + B_y^2 +B_z^2}$ & 0.3068 & 0.1160 & 0.2788 & 0.1342 & 0.1212 & \textcolor{red}{\textbf{0.1065}} \\
         III.17.37 & $\beta ( 1+ \alpha cos\theta )$ & 0.0962 & 0.0487 & 0.3066 & 0.0372 & \textcolor{red}{\textbf{0.0191}} & 0.0299 \\
         \hline
    \end{tabular}
    }
    \caption{RMSE Comparison on Function Approximation Task of Feynman Dataset}
    \label{tab:RMSE Comparison on Funtion Fitting}
\end{table*}

\section{Experiment}
\subsection{Dataset and Implementation Setting}
The best performance is reported as the average of three independent runs using the random seed "2024". It is worth noting that this seed is used solely for reproducibility and reflects the year of the paper's writing.

\textbf{Dataset}: We conduct our experiments on benchmarking datasets spanning image, text, multimodal and function approximation. Specifically, we evaluate our proposed method on the following datasets: MNIST\cite{mnistdatabase}, CIFAR10\cite{krizhevsky2009learning}, CIFAR100\cite{krizhevsky2009learning}, STL10\cite{coates2011analysis}, AG News\cite{zhang2016characterlevel}, and AVMNIST\cite{liang2021multibench}, MIMIC-III \cite{Harutyunyan2019,johnson2016mimic}, Feynman \cite{liu2024kan,udrescu2020ai}. Among these, AG News is a sentiment analysis dataset, while AVMNIST is a multimodal dataset comprising image and audio spectrum modalities of handwritten numbers and MIMIC-III is a multimodal dataset including patient ICU monitoring records and electronic health records compromising irregular time series and text modalities. Feynman, on the other hand, is a symbolic regression and function approximation dataset of physical equations collected from Feynman's textbook. We implement a function approximation task using the Feynman dataset, as suggested by Liu et al.\cite{liu2024kan}, to investigate whether FR-KAN can learn better activation functions compared to KAN. We use the pruned KAN architecture recommended by Liu et al.\cite{liu2024kan} and set $G=20$. Synthetic data for the function approximation task is generated using pykan. However, some functions in the dataset occasionally generate NaN input or Inf values in the labels, and these instances are excluded from our experiments.

\textbf{Baseline}: We select rectified MLP, original KAN\cite{liu2024kan}, Fourier-KAN\cite{xu2024fourierkan}, Rational-KAN\cite{yang2024kolmogorov} and RBF-KAN\cite{li2024kolmogorovarnold} as our baseline. Fourier-KAN, Rational-KAN and RBF-KAN replaced B-spline function with their corresponding basis function. We control the number of trainable parameters to build the model architecture. For FR-KAN and MLP, hidden width is in selected from [32, 64, 128, 256, 512, 1024] and the depth of layers ranges from 1 to 3. For original KAN, we use the hidden neuron from [4, 8, 16, 32, 64, 128] to control the trainable parameters closed to FR-KAN and MLP. Moreover, we control the grid size of KAN and proposed method as 20 since we empirically find this is the sweet spot to balance performance and efficiency. For all baseline method, a layernorm layer is added before each KAN-Linear layer to stabilize the training.

\begin{figure*}
    \centering
    \includegraphics[width=0.95\linewidth]{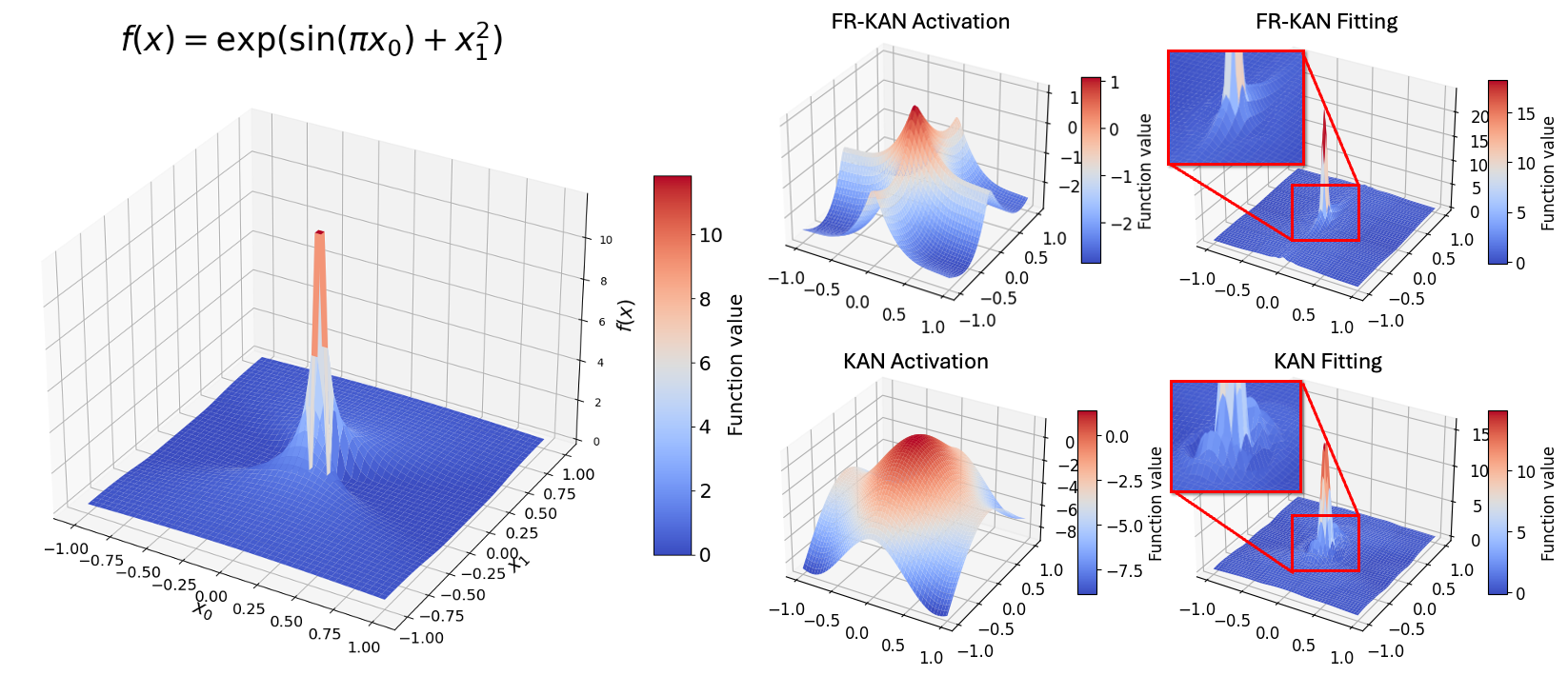}
    \caption{Visualization on Complex Learned Activation and Function Approximation of KAN and FR-KAN}
    \label{fig:Visual_Fitting}
\end{figure*}

\subsection{Classic Deep Learning Task}
We shows the Performance-Parameter plots in Figure \ref{fig:All_performance} and select ReLU-MLP, KAN, and two typical and strongest KAN variants, Rational-KAN and RBF-KAN as the baseline model. We exclude Fourier-KAN in Figure \ref{fig:All_performance} since the incomparable Performance-Parameter ratio. Adding Fourier-KAN to Figure \ref{fig:All_performance} will make the plots difficult on presentation.

\textbf{Control Parameter}: Row 1 of Figure \ref{fig:All_performance} displayed model performance on image dataset, where we compare the performance with trainable Parameter at the same time. It is clearly to see that FR-KAN can maintain the same level as MLP across all image dataset while KAN's performance is limited because of the trainable parameter. In MNSIT, FR-KAN achieve the same performance as MLP, but the complexity of MNIST limit the performance. When the dataset goes complex such as CIFAR100 and STL10, we observed a clear performance increase over MLP. The best performance in CIFAR10, CIFA100, STL10 are achieved by FR-KAN at 58\%, 30\% and 48\%. Although performance of KAN increase with trainable parameter but the efficiency is still less competitive to MLP and FR-KAN due to unique learnable activation requires $(G+K)$ times more complex than MLP. Moreover, original KAN displays a higher performance variance compared to FR-KAN and MLP, reflecting stabler training than KAN.

Moreover, we further compare all the model in multimodal, NLP, and time series scenario. We find that FR-KAN also generally performed better than others. Interestingly, KAN-based network always perform better than ReLU-MLP in multimodal setting and time series and shows relatively good stabilization proven by the small performance variance. This may be a positive signal that KAN-based method is good at handling complicated task than MLP. It is interesting that MIMIC-III dataset has the biggest performance variance compared to other dataset. We think the reason is that the dataset is collected from real-world clinical scenario, which may include many noisy sample. In NLP scenario, original KAN shows very unstable performance than others while FR-KAN has the smallest variance and highest performance.

\textbf{Control Hidden Layer}: We further controlled the hidden structure across models and evaluated their performance. As shown in Table \ref{tab:Comparison_Hidden_Structure}, FR-KAN consistently outperformed the original KAN and MLP on the CIFAR10 dataset, and lost in only one configuration on the STL10 dataset. Additionally, we observed that increasing the number of neurons per layer is significantly more effective than stacking additional layers for KAN-based methods. This aligns with the intuition from the Kolmogorov-Arnold Theorem, as a higher number of neurons enables the generation of more univariate functions, enhancing the model capacity and expressive power.

\subsection{Function Approximation}
Apart from classic deep learning task, we also compare proposed method and baselines in function approximation as stated in Table. \ref{tab:RMSE Comparison on Funtion Fitting}. We use the function generation script offered by pykan\cite{liu2024kan} and evaluate model performance on Feynman dataset. We exclude some equation suggested by Liu et.al \cite{liu2024kan} since the function generation script returns NaN in input and inf in label. However, the remaining equations are still representative to evaluate the propose method. To be fair, we control the model depth and width by adopting the "Pruned KAN" setting suggested in \cite{liu2024kan}. It is clear that our FR-KAN wins over most functions in terms of RMSE. MLP is not comparable to KAN-based architecture in this experiments, reflecting the strong function approximation ability of KAN-based methods. Rational-KAN and RBF-KAN performs closely in RMSE. Original KAN also wins many than Rational and RBF KAN in many function, showing that spline-based KAN has stronger expressive power.

\section{Can FR-KAN Find Better Activation than KAN?}
The interpretability of KAN come from Kolmogorov-Arnold Theorem, which a complex function can be expressed as a superposition of continuous single variable functions. Each of neuron in the network can be considered as a univariate function. A good univariate function candidate should be similar to final prediction and fit well to target function. To show the advantage and stronger interpretability of FR-KAN, We take function I.6.2 of Table \ref{tab:RMSE Comparison on Funtion Fitting} containing two input variables and visualize the learned activation and function approximation of KAN and FR-KAN in Figure \ref{fig:Visual_Fitting}. We set the grid size as 10 and order as 3 for this visualization. I.6.2 function is displayed on the left hand size where the bottom of the function is smooth and slim column in the middle. We can see that KAN learned a relatively round activation looks like a bell while FR-KAN is sharp in the middle part and smooth at the bottom. FR-KAN fits a more similar activation than KAN. On the other hand, we find that under the same condition, KAN generate serious oscillation at the bottom of function, making it diverge from the actual function value. While FR-KAN can fit a much smoother without oscillation, indicating the expressive power of FR-KAN is actually better than KAN due to the free knots shifts.

\begin{figure}
    \centering
    \includegraphics[width=0.95\linewidth]{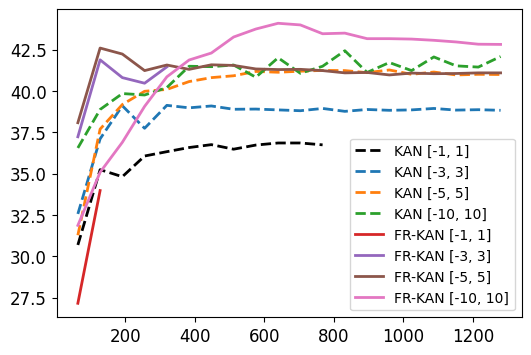}
    \caption{Large Grid Range to Stabilize Training: Accuracy vs Training Step on STL10 Dataset for more than 1200 steps. KAN [-1, 1] and FR-KAN[-1, 1] stop early due to NaN loss}
    \label{fig:Impact_Grid_Range}
\end{figure}

\section{Larger Grid Range for Better Activation Field}
Unlike the original KAN and its variants, which set the grid range to [-1, 1] based on the normal distribution assumption in deep learning models, we found that such a restricted range can severely hinder model convergence. From Theorem \ref{theorem:KAN_Knots_Bound}, it is evident that the number of knots is highly dependent on the grid size. However, simply increasing the grid size $G$, such as through grid extension in the original KAN, does not necessarily improve performance for typical deep learning tasks. This is because the activation function can become overly complex and zigzagged, leading to rapidly varying gradients that are difficult to optimize. To overcome this limitation, FR-KAN adopts a larger grid range (e.g, [-10, 10]) instead of the traditional narrow range. The impact of using a larger grid range is demonstrated in Figure \ref{fig:Impact_Grid_Range} with incremental grid range. This experiment excludes the impacts of SiLU of original KAN stated in Equation \ref{eq:KAN_architecture} but only remains $A_b \sum_{j=0}^{G} c_j B_j(x_i)$. As shown from Figure \ref{fig:Impact_Grid_Range}, when grid range set to [-1, 1], KAN and FR-KAN all meet NaN training accuracy since the limited grid range. As the grid range get extend to [-3, 3], we can clearly see that the models are easier to converge with higher accuracy without NaN Loss. With the increasing grid range, we can see the that the performance raise to higher value and the FR-KAN indicate better training stability compared to KAN.

\section{Conclusion and Future Works}
In this work, we derive the knots upper bound for B-spline KAN and Free knots upper bound for FR-KAN. We also validate the propose method in various modality dataset and show that FR-KAN is less parameterized than original KAN. The experiments demonstrate that our KAN model can not only perform better than MLP and original KAN in terms of classic deep learning task but also wins on function approximation task due to the smooth regularizer and free knots features. 

However, there are still some limitation on our works. Firstly, the backward propagation usually introduce more gradient since the grids parameter needs to be updated. Therefore, further optimization on code implementation is required to implement more efficiently from CUDA. High computational graph is the shortage of B-spline function during the backward phrase due to recursive generation of activation. The future research can focus on reducing computational graph from recursive round of B-spline.

\bibliography{main}
\bibliographystyle{icml2025}


\newpage
\appendix
\onecolumn
\section{Proof Lemma \ref{lemma:knots_mul_layer}}
\label{proof:Lemma_kots_mul_layer}

Next, let us prove Lemma \ref{lemma:knots_mul_layer} the knot upper bound of ReLU network. Given an arbitrary initialized fully connected Multi-layer Perceptron (MLP) with ReLU activation function is a composition of layers of computational units which maps the input dimension $q$ to output dimension $p$, $F_r^l: \mathbb{R}^{q} \rightarrow \mathbb{R}^{p}$, the linear transformation in $F_r^l$ is an affine transformation that generates a linear spline decision boundary without lost of knots.

\begin{equation}
    F(x, w, b) = \sum^{n}_{j=0}w_{l, j} \sigma(w_{l-1, j} x + b_{l-1, j}) + b_{l, j}
\end{equation}

This can simply equal to 

\begin{equation}
    F(x, w, b) = \sum^{n}_{j=0}w_{l, j} w_{l-1, j}  \sigma(x + \frac{b_{l-1, j}}{w_{l-1, j}}) + b_{l, j}
\end{equation}

Since the $\sigma=max(x, 0)$ is ReLU activation function. It is naturally to see that this MLP is linearly spline function with knots point located at $t_j=\frac{b_{l-1, j}}{w_{l-1, j}}$. For simplicity, we denote the point as $t_j=\frac{b_{j}}{w_{j}}$. Therefore, we can define that the knot is unique if the ratio of bias and weight are unique. Naturally, a spline function from previous layer $l-1$ can be preserve in affine transformation in next layer $l$. In addition, if the affine transformation maps the linear spline to negative area of ReLU activation, then those knots in negative area will be destroyed. Then, we know that facts that any neuron $n_i \in F_r^l$ can preserve at most $m_{i-1}$ knots from previous layer. If the affine transformation maps the linear spline to positive area and $t_{j}$ is unique. Then one new knot will be added to the spline. Then, the upper bound of spline knots for MLP can be obtained when $wx+b$ are all positive and $t_{j}$ is unique as stated in Lemma \ref{lemma:knots_mul_layer}.

\begin{equation}
    m_l \leq m_{l-1} + n_i(m_{l-1}+1)
\end{equation}

\section{Proof of Theorem \ref{theorem:KAN_Knots_Bound} and Tightness of the Bound}
\label{proof:tightness_proof}
From Lemma \ref{lemma:knots_mul_layer}, we understand that the neuron can preserve the knots from previous layer. Following this idea, we can derive the knots upper bound of KAN. Let us denote $\triangle y_i = y_{i-1} - y_{i}$ as the difference of activation output in the grid interval, $\triangle g$ as the grid segment and $K_b^l$ as the fixed grid KAN network. We start from the first layer of KAN $K_b^1$, the knots of first layer only generate knots equivalent to $(G+K)$ as all the neuron share the same grid. To introduce more knots, we make $K_b^l$ deeper such that $l > 1$. The knots can be generated only if $\triangle y_i > \triangle g$. For each grid segment, one can generate unique knots at most $G-1$ since the two knots at grid range $[a,b]$ are overlapped with previous layer. For each neuron, there are $G$ grid segment in $[a, b]$. To reach the maximum new knots in segment $g_i$, we can assume the KAN can obtain $\lfloor \frac{\Delta y_i}{\Delta g_i} \rfloor$ additional knots for each segment. Then, we can obtain the knots upper bound of KAN with B-spline function with $G$ grid size and $K$-order as follows.

\begin{equation}
    (G+K) \leq \mathcal{N}_k \leq (G+K) + \prod_{l=1}^l G(G-1)
\end{equation}

\begin{equation}
    \mathcal{N}_k = m_{l-1} + \sum_{i=1}^G \lfloor \frac{\Delta y_i}{\Delta g_i} \rfloor
\end{equation}

It is worthy to note that this bound is tight. This can be proved by constructing a series of $c_j$ that reach the upper bound. Next, we show this bound is tight for any $K_b^l$. One way to prove this tight bound is to construct a sawtooth activation after linear combination. Given a fixed knots KAN network with B-spline function $K_b^l$ expressed as Equation \ref{eq:KAN_architecture}. For simplicity, we set the B-spline order as $K=1$, $A_b$ is an identity matrix and remove the $SiLU$ shortcut since it does not contribute to number of knots. We rewrite the KAN network as Equation \ref{eq:simple_KAN}. Then, there exist a spline combination parameter $c_j$ as stated in Equation \ref{eq:special_spline_weight} :

\begin{equation}
    \phi(x_i) = I \sum_{j=0}^{G} c_j B_j(x_i) \label{eq:simple_KAN}
\end{equation}

\begin{equation}
B_{j,0}(g) =
    \left\{
    \begin{aligned}
        0, & & & \text{if } g_j \leq g < g_{j+1}, \\
        1, & & & \text{otherwise.}
    \end{aligned}
    \right.
\end{equation}

\begin{equation}
    \label{eq:b_spline_basis_fn}
    B_{j, k}(g) = \frac{g - g_j}{g_{j+k} - g_j}B_{j, k-1}(g_j) + \frac{g_{j+k+1} - g}{g_{j+k+1} - g_{j+1}} B_{j+1, k-1}(g)
\end{equation}

\begin{equation}
    \label{eq:special_spline_weight}
     c_j = (-1)^j, \ \  j=1,2,\dots,G
\end{equation}

\begin{figure*}
    \centering
    \includegraphics[width=0.9\linewidth]{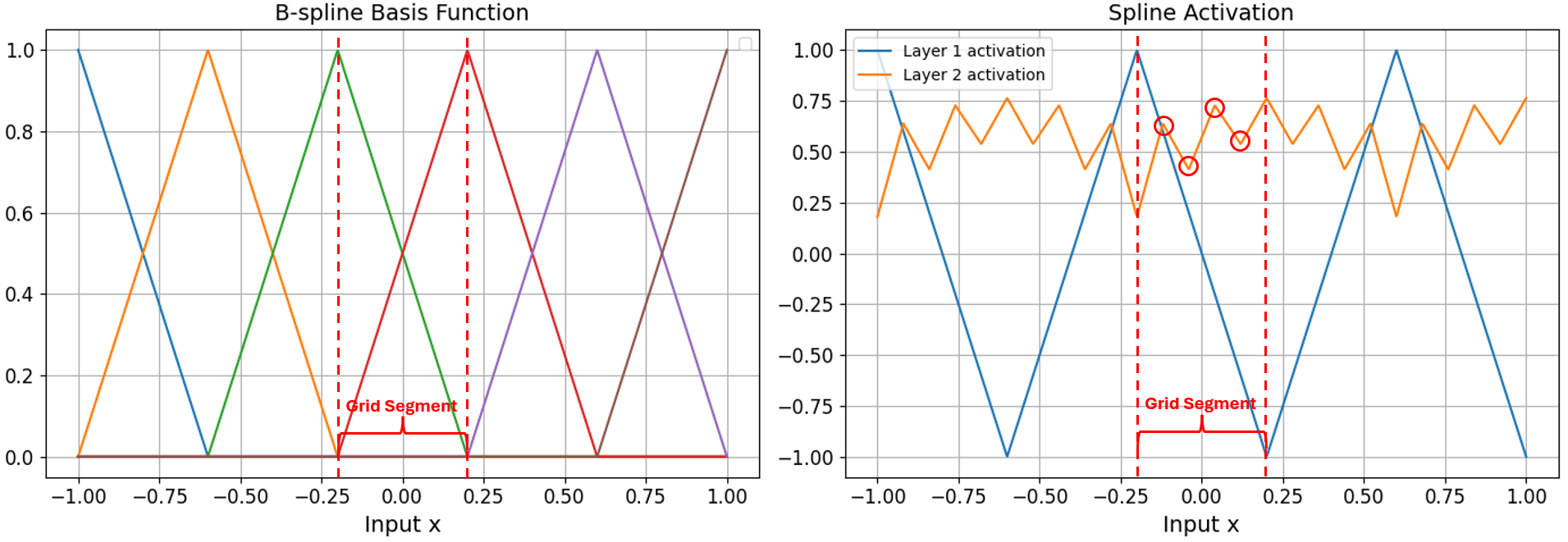}
    \caption{A spline activation example, where $K=1$ and $G=5$. \textbf{Left}: Basis function of B-spline $B_{j,k}$. \textbf{Right}: Spline activation function $\sum_{j=0}^{G} c_j B_j(x_i)$. We highlight the new knots in layer 2 activation with \textbf{\textcolor{red}{\(\bigcirc\)}}.}
    \label{fig:spline Example}
\end{figure*}

Given a B-spline basis function stated Equation \ref{eq:b_spline_basis_fn} with degree $k$, the partition of unity stated that $\sum_{j=0} B_{j, k} = 1$. In addition, the order of B-spline determines the number of activated basis $B_{j, k}$ in a grid segment $g_i$. Therefore, our constructed example above can give us an activation span the entire range $[-1, 1]$ as shown in Figure \ref{fig:spline Example}. We construct a layer 1 activation spline that spans the entire grid range, which fulfill the condition $\triangle y_i \geq G\triangle g$. Therefore, layer 2 activation introduce $G-1$ new knots, where the spline weight $c_{j,k}$ in layer 2 are randomly initialized from $[0,1]$. Those new knots introduce more oscillation. This derivation can be naturally extend to higher order spline function, which concludes the upper bound of our Theorem \ref{theorem:KAN_Knots_Bound} can be reached, which is a tight bound.

\end{document}